\documentclass[11pt]{article}

\usepackage[final]{acl}
\usepackage{times}
\usepackage{latexsym}

\usepackage[T1]{fontenc}

\usepackage[utf8]{inputenc}
\usepackage{bm} 

\usepackage{microtype}

\usepackage{inconsolata}

\usepackage{graphicx}

\usepackage{hyperref}       
\usepackage{url}            
\usepackage{booktabs}       
\usepackage{amsfonts}       
\usepackage{nicefrac}       
\usepackage{microtype}      
\usepackage{xcolor}         
\usepackage{listings}
\usepackage{comment}
\usepackage{amsmath}
\usepackage{multirow}
\usepackage{adjustbox}
\usepackage[table]{xcolor}
\usepackage{threeparttable}
\usepackage{graphicx}
\usepackage{caption}
\usepackage{subcaption}
\usepackage{enumitem}
\usepackage[most]{tcolorbox}
\usepackage{lipsum} 
\usepackage{geometry}
\tcbuselibrary{breakable}
\usepackage{titlesec}
\usepackage{dblfloatfix}
\usepackage{soul}
\usepackage[ruled,vlined]{algorithm2e}
\usepackage{mathrsfs}

\definecolor{highlightcolor}{gray}{0.90}
\lstdefinestyle{sparql}{
  language=SQL, 
  basicstyle=\ttfamily\small,
  keywordstyle=\color{blue},
  stringstyle=\color{red},
  commentstyle=\color{gray},
  morekeywords={SELECT, WHERE}, 
  frame=single,
  breaklines=true,
  columns=fullflexible
}

\title{KG-Reasoner: A Reinforced Model for End-to-End Multi-Hop Knowledge Graph Reasoning}
\author{
  Shuai Wang \quad \quad \quad \quad \quad Yinan Yu \\
  Department of Computer Science and Engineering \\
  Chalmers University of Technology and University of Gothenburg \\
  SE-41296 Gothenburg, Sweden \\
  \texttt{\{shuaiwa, yinan\}@chalmers.se}
}

\begin{document}
\maketitle

\begin{abstract}
Large Language Models (LLMs) exhibit strong abilities in natural language understanding and generation, yet they struggle with knowledge-intensive reasoning. Structured Knowledge Graphs (KGs) provide an effective form of external knowledge representation and have been widely used to enhance performance in classical Knowledge Base Question Answering (KBQA) tasks. However, performing precise multi-hop reasoning over KGs for complex queries remains highly challenging.
Most existing approaches decompose the reasoning process into a sequence of isolated steps executed through a fixed pipeline. While effective to some extent, such designs constrain reasoning flexibility and fragment the overall decision process, often leading to incoherence and the loss of critical intermediate information from earlier steps. In this paper, we introduce KG-Reasoner, an end-to-end framework that integrates multi-step reasoning into a unified "thinking" phase of a Reasoning LLM. Through Reinforcement Learning (RL), the LLM is trained to internalize the KG traversal process, enabling it to dynamically explore reasoning paths, and perform backtracking when necessary.
Experiments on eight multi-hop and knowledge-intensive reasoning benchmarks demonstrate that KG-Reasoner achieves competitive or superior performance compared to the state-of-the-art methods. \footnote{Codes are available at the repository: \url{https://github.com/Wangshuaiia/KG-Reasoner}.} \looseness=-1
\end{abstract}

\section{Introduction}

\begin{figure}[t]
    \centering
    \includegraphics[width=\linewidth]{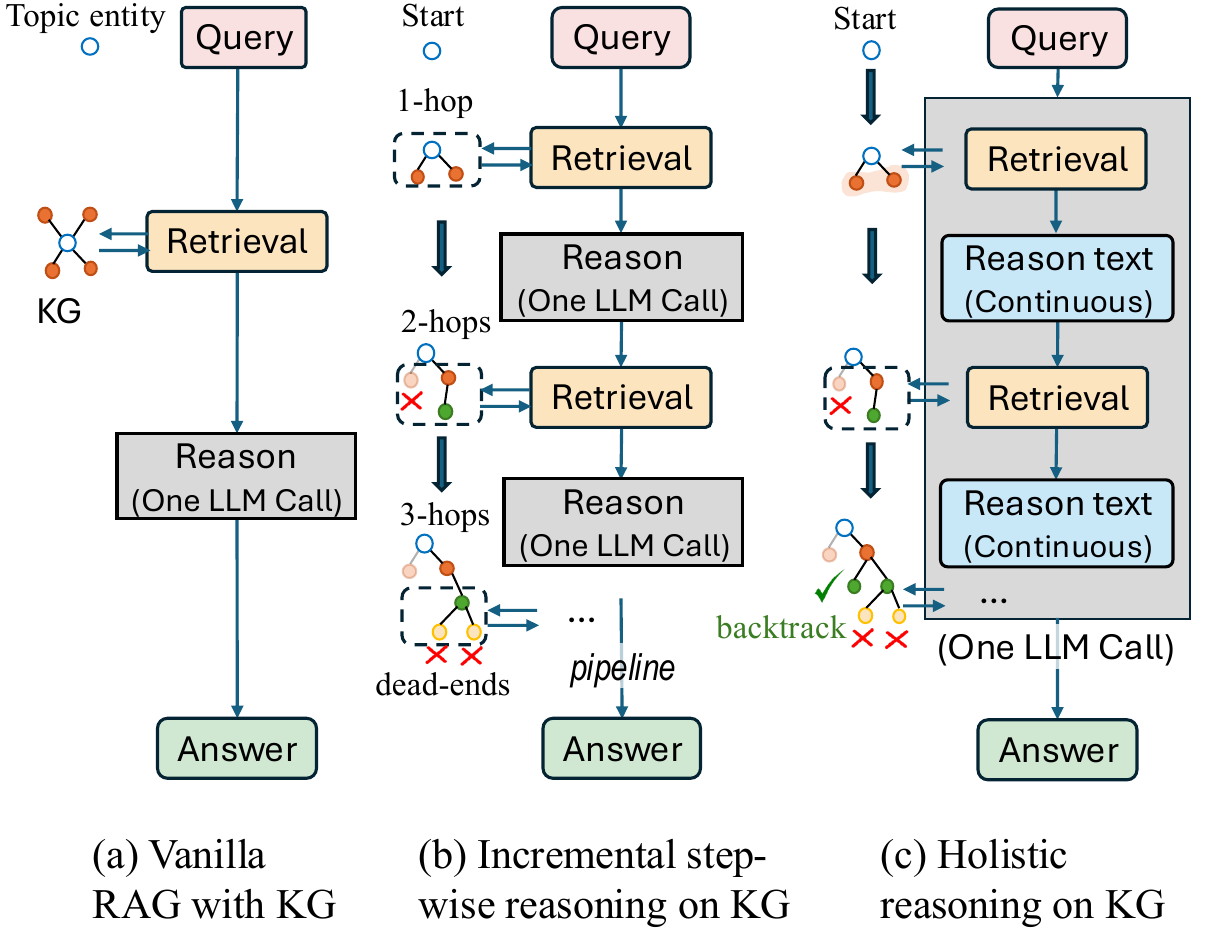}
    \caption{Comparison of KG-based reasoning paradigms. (a) and (b) illustrate two representative pipeline-style approaches that leverage knowledge graphs through staged retrieval and reasoning. (c) shows our proposed end-to-end holistic reasoning framework, which integrates retrieval and multi-hop reasoning within a unified thinking process.}
    \label{fig:compare-intro}
\vspace{-0.5cm}
\end{figure}

Large Language Models (LLMs) have achieved remarkable success across diverse domains but still exhibit notable shortcomings, such as hallucinations and factual inaccuracies, particularly in knowledge-intensive tasks \cite{agrawal2024can}. This stems from the implicit encoding of knowledge within model weights, making knowledge updates cumbersome and resource-intensive when relying on fine-tuning. To address this limitation, Retrieval-Augmented Generation (RAG) has emerged as an effective strategy, enabling LLMs to dynamically access external knowledge during inference~\cite{gao2023retrieval}. 
Compared with textual data, knowledge graphs (KGs) provide structured and explicit relations between entities, allowing LLM reasoners to ``hop'' across connected entities and discover links that traditional semantic retrieval may miss. This graph-based nature significantly broadens the retrievable knowledge domain and enhances reasoning potential.

A direct approach to leveraging KGs involves prompting LLMs to generate structured queries, such as SPARQL \cite{li2024flexkbqa,zahera2024generating}, which are then executed to retrieve relevant information, as illustrated in Figure \ref{fig:compare-intro}(a). To better incorporate contextual information, some methods retrieve subgraphs and prune them to keep only relevant information~\cite{hu-etal-2025-grag,xu2024retrieval}. GraphRAG \cite{edge2024local} further improves retrieval accuracy and coverage on large-scale KGs by partitioning the graph into topic-centric subgraphs via community detection and performing retrieval within each subgraph. 
Despite their effectiveness, these retrieval-based methods mainly handle low-hop questions, where answers are close to topic entities. In contrast, many real-world queries require multi-hop reasoning, involving intermediate inferences across distant entities. For example, answering “\textit{What is the official flower of the area affected by Tropical Storm Fabio?}” requires first identifying the affected area and then retrieving its official flower. In such cases, one-shot retrieval and prompting are often insufficient.

To handle multi-hop reasoning over KGs, typical methods first identify entities in the question, then progressively expand along graph edges to collect relevant paths. Such pipeline is shown in Figure \ref{fig:compare-intro}(b). For example, KG-CoT \cite{zhao2024kg} models reasoning as a stepwise transition process with a learned transition matrix. More recent LLM-based approaches treat the KG as an interactive environment, where an LLM-based agent performs iterative reasoning and decides at each step which neighboring entity to explore next~\cite{sun2024thinkongraph,xiong-etal-2024-interactive}. Building on this paradigm, additional techniques such as memory mechanisms~\cite{jiang-etal-2025-kg,chen2024plan} or Monte Carlo Tree Search (MCTS)~\cite{luokbqa} have been incorporated to enhance the exploration of reasoning paths.

However, most existing multi-hop reasoning pipelines invoke LLMs separately at each reasoning step. Intuitively, this design is suboptimal, as it fragments the reasoning process into a sequence of isolated decisions, whereas answering complex questions ideally requires continuous and coherent reasoning.
Moreover, errors introduced at early stages tend to propagate and amplify in downstream steps.
Although error detection and backtracking could mitigate this issue in principle, such mechanisms are difficult to implement effectively under the current pipeline-style setting, where intermediate reasoning steps are executed independently.

To address these limitations, we seek a unified approach that integrates reasoning and retrieval within a single framework. Unlike existing pipelined approaches, we consider using a single LLM to perform multiple reasoning and retrieval steps in an end-to-end manner, as shown in Figure \ref{fig:compare-intro}(c). Inspired by recent advancements in \textit{Reasoning LLMs}, such as \textit{GPT-5 thinking} \cite{openai_gpt5} and \textit{DeepSeek-R1}~\cite{guo2025deepseek}, these models prioritize an extended internal reasoning phase before generating a final response. Through reinforcement learning (RL) and self-play, the model acquires advanced cognitive abilities, including backtracking from incorrect paths and verifying intermediate steps. These capabilities are precisely what multi-hop reasoning requires.  We therefore embed both retrieval and reasoning operations directly into a unified long-thinking process, preserving reasoning continuity while enabling efficient multi-step inference.

In this paper, we propose a reinforcement-based model for end-to-end multi-hop knowledge graph reasoning. Retrieval calls and multi-hop reasoning are jointly conducted within a single LLM thinking process. We introduce a graph-based backtracking mechanism that enables the model to revisit historical reasoning paths and reselect entities upon error detection. Additionally, we employ a graph attention network for entity selection, allowing the model to aggregate richer contextual information from the KG and improve selection accuracy. To further enhance multi-hop reasoning over the KG, we train the model using RL with carefully designed reward functions. 
Our contributions are summarized as follows:

\begin{itemize}[leftmargin=1.5em]
\item We propose an end-to-end framework for multi-hop KBQA that unifies retrieval and reasoning within a single thinking phase of an LLM call, enabling coherent multi-step reasoning.

\item We enhance multi-hop reasoning via reinforcement learning with tailored rewards, GNN-based entity selection, and explicit error backtracking mechanism.

\item We conduct extensive experiments on eight multi-hop and knowledge-intensive reasoning benchmarks, demonstrating the effectiveness of the proposed approach.

\end{itemize}

\section{Related Work}
\label{sec:related-work}
\subsection{Knowledge Graph Reasoning}
Knowledge Base Question Answering (KBQA) is a prevalent task aimed at reasoning over a KG to find answers to given questions. Traditional KBQA approaches are mainly categorized into Information Retrieval (IR)-based and Semantic Parsing (SP)-based methods. IR-based methods retrieve relevant entities and relations from the KG to infer answers, while SP-based methods translate questions into executable queries (e.g., SPARQL) that are directly executed on the KG~\cite{saxena2020improving, dai2023fedgamma, chen2022temporal}.

With the advent of LLMs, recent studies have leveraged these models to enhance KG reasoning by decomposing complex questions into sequential sub-questions~\cite{ma2025debate, jiang-etal-2025-kg, ao2025lightprof, zhang2025collaborative, sun2024thinkongraph}. 
Owing to the structured nature of KGs, RL has been widely used to guide reasoning path exploration. Prior RL-based approaches typically adopt RNN agents to iteratively predict paths~\cite{das2018go, lin2018multi, cui2023incorporating,wang2026kg}, but often suffer from inefficient exploration in multi-hop reasoning. To address this, Zhang et al.~\cite{zhang2025collaborative} proposed a collaborative framework combining LLM priors with RL.
However, existing approaches still rely on incremental, separate reasoning steps, leading to limited context retention and potential error propagation.

\subsection{Reasoning LLMs}
Recent progress demonstrates that explicitly prompting LLMs to perform thinking before give an answer significantly enhances their problem-solving capabilities~\cite{li2025system, wuthinking}. Reinforcement Learning has further strengthened these abilities by enabling LLMs to iteratively reflect and reason effectively~\cite{wang2024math, zhu2025chain}, as evidenced by \textit{Reasoning LLMs} such as Deepseek-R1~\cite{guo2025deepseek}, Kimi-k1.5~\cite{team2025kimi} and GPT-5 thinking~\cite{openai_gpt5}. RL has also facilitated the integration of LLMs with external retrieval tools, particularly in multi-hop RAG frameworks~\cite{song2025r1, li2025webthinker, chen2025research, sun2025zerosearch}, which typically involve iterative searches without strict structural constraints. 
In contrast, multi-step reasoning over KGs poses greater challenges, since paths must be explored within a rigid, structured space. To address this, we place the reasoning process inside the LLM’s “thinking” phase, thereby leveraging the capabilities of reasoning LLMs to perform efficient end-to-end KG reasoning.

\begin{figure}[t]
    \centering
    \includegraphics[width=\linewidth]{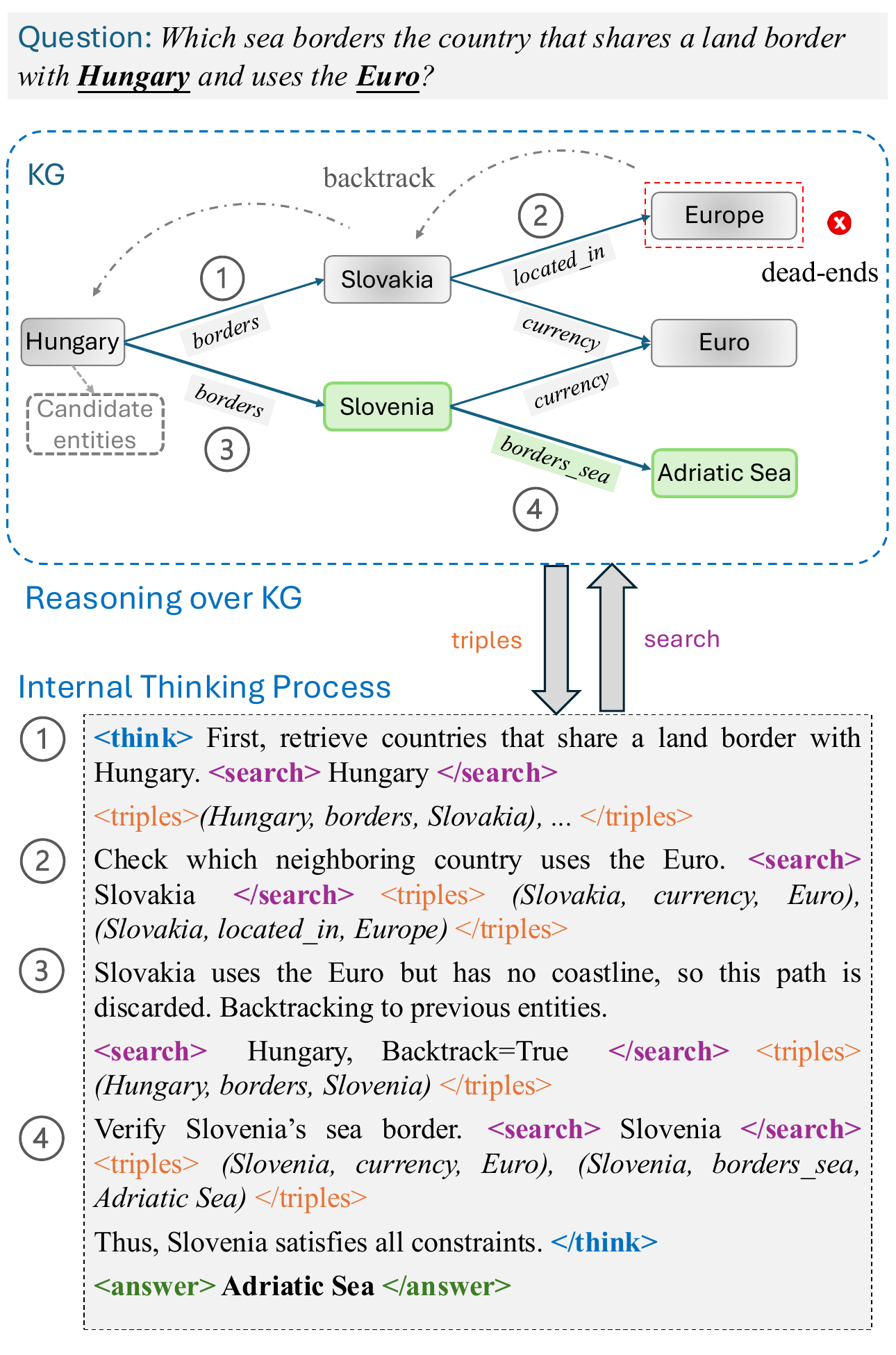}
    \caption{The architecture of our end-to-end reasoning framework.}
    \label{fig:main}
\end{figure}

\section{Task Definition}
We address the task of Knowledge Base Question Answering (KBQA) over a knowledge graph $G = \{(e, r, e') \mid e, e' \in \mathcal{E}, r \in \mathcal{R} \}$, 
where \( \mathcal{E} \) denotes the set of entities and \( \mathcal{R} \) denotes the set of relations.
Each triple \( (e, r, e') \) represents a factual relation \( r \) between two distinct entities \( e \) and \( e' \).
Given a natural language question $q$, along with an identified topic entity $e_t \in \mathcal{E}$ mentioned in the question, the objective is to infer the correct answer entity $e_a \in \mathcal{E}$ by reasoning over the knowledge graph $G$.
The answering process typically starts from the topic entity $e_t$ and involves exploring the graph to identify the answer entity $e_a$. In practice, large-scale knowledge graphs are often sparse and incomplete, making it difficult to locate the answer through simple one-hop queries. As a result, multi-hop reasoning is frequently required, where the answer entity $e_a$ may reside several hops away from the topic entity $e_t$.

\section{Method}

The overall framework of our method is illustrated in Figure \ref{fig:main}. Our approach builds upon a reasoning-oriented large language model (LLM) that is capable of performing multi-hop search and inference in an end-to-end manner. During the reasoning phase, the LLM autonomously invokes the search tool multiple times and evaluates the correctness of the intermediate reasoning after each search step. Once sufficient evidence has been collected and the reasoning process is completed, the LLM transitions to the answering phase to generate the final answer. The reasoning workflow of our approach is summarized in Algorithm 1.

\subsection{Retrieval Calling}
\label{Retrieval-method}

To enable the LLM to interact with external tools, we follow a similar approach to previous studies~\cite{song2025r1, chen2025research}, where the model generates specific tokens to invoke tools. Specifically, we define special tokens \texttt{<search>} and \texttt{</search>} that, when generated by the LLM, automatically trigger a custom retrieval tool. The retrieved results are then wrapped with a special tag \texttt{<triples>}, allowing the model to continue reasoning or perform further searches based on the retrieved content. To help the LLM learn to generate such tokens and adapt to such reasoning pattern, we fine-tune it using a dataset formatted in this format. Below is an example of the data format. Given the question:
\textit{“What timezone is Utah in?”}
The simplified response is:

\textit{\textcolor{blue}{<think>}Given the information, I don't have specific knowledge about Utah's timezone. I will perform a search for the required information. The query to search for is \textcolor{purple}{<search>}utah\textcolor{purple}{</search>} \textcolor{orange}{<triples>} (Utah, timeZone, Mountain Time Zone)\textcolor{orange}{<triples>}. Based on the search results, I found that Utah is in the Mountain Time Zone.[...]\textcolor{blue}{</think>} \textcolor{teal}{<answer>}Mountain Standard Time\textcolor{teal}{</answer>}}

After each retrieval step, the LLM continuously assesses the relevance of the retrieved content and correctness of the reasoning process. If the LLM determines that the information is unrelated or incorrect, the custom KG retrieval tool is called again with the parameter \texttt{Backtrack=True}. This reasoning process continues until sufficient information has been gathered.

Details of the custom KG retrieval tool are presented in Section~4.2, and the construction of the fine-tuning dataset is provided in the Appendix.


\begin{algorithm}[t]
\caption{Reasoning with LLM over Knowledge Graph}
\label{alg:reasoning-llm}
\KwIn{Question $q$ with topic entity $e_{\text{topic}}$, Knowledge Graph $\mathcal{G}$}
\KwOut{Answer $a$ to question $q$}

Initialize LLM input with $q$\;
Start with \texttt{<think>} to begin reasoning\;

\While{not reached \texttt{<answer>} stage}{
    LLM analyzes current context and generates query entity $e_q$ within \texttt{<search>} and \texttt{</search>} tags\;
    
    \If{\texttt{</search>} is emitted}{
        Execute retrieval over $\mathcal{G}$ using $e_q$\;
        Obtain result triples $\mathcal{T}$\;
        Wrap $\mathcal{T}$ in \texttt{<searched\_triples>} and feed back to LLM\;
    }

    // LLM reflects on current entities and triples\;
    \If{retrieved information is sufficient}{
        End reasoning with \texttt{</think>}\;
        LLM proceeds to generate 
        final answer within \texttt{<answer>} and \texttt{</answer>}\;
    }
    \Else{
        LLM decides whether to continue exploring or revise query\;
    }
}

Extract answer $a$ from the \texttt{<answer>} tag\;
\Return $a$\;

\end{algorithm}

\subsection{Knowledge Graph Retrieval Tool}
\label{gnn-method}

Large-scale knowledge graphs are typically stored in graph databases and queried using SPARQL for retrieval. Our retrieval process starts from a topic entity and expands the search on the KG.

\subsubsection{One-Hop Neighbor Retrieval}

Given a topic entity, we use SPARQL to retrieve all predicates and corresponding object entities linked to it from data bases. For example, given the question \textit{``What books did J.K. Rowling write?''}, the topic entity is \texttt{J.K. Rowling}. The following SPARQL query is issued to retrieve all outgoing predicates and their objects:

\begin{lstlisting}[style=sparql]
SELECT ?predicate ?object
WHERE {
  ns:m.05b6w ?predicate ?object .
}
\end{lstlisting}

\noindent
Here, \verb|ns:m.05b6w| denotes the Freebase ID for J.K. Rowling, \verb|?predicate| retrieves all relations where she appears as the subject, and \verb|?object| returns the corresponding object entities. 
Similarly, we query incoming predicates where the entity appears as the object. 

\subsubsection{GNN-based Entity Ranking}
\label{sec:gnn}

Since not all entities in the knowledge graph are relevant to the question, we perform entity ranking to score their relevance and select the top-$k$ entities as candidates. The semantic information carried by an entity alone is often limited; however, its one-hop neighbors can provide rich contextual information. For instance, an entity’s attributes are often represented as one-hop neighbors in the graph.

To effectively incorporate such contextual signals, we adopt a graph neural network (GNN) that aggregates information from an entity’s neighbors. We employ an attention mechanism to selectively focus on the neighbors most relevant to the question. Given an entity $h$ with a set of one-hop neighbors $(h_i, t_i)$ connected via relation $r_i$, the enhanced representation $\bm{\hat{h}}$ is computed as:

\begin{equation}
\begin{aligned}
    s_{i} &= {a} (\mathbf W \bm q;\, \mathbf W (\bm t_i\odot\bm r_i))\\
    \alpha_i &= {\rm softmax}(s_i)\\
    \bm{\hat{h}} &= \sigma \Big(\mathbf W_h [\bm{h} ~\Vert \sum_{(h_i, t_i)\in\mathcal N(h)} \alpha_i\, \bm t_i]\Big)
\end{aligned}
\end{equation}

Here, $a: \mathbb{R}^{d} \times \mathbb{R}^{d} \rightarrow \mathbb{R}$ is a function that computes attention scores between the question $\bm{q}$ and each neighbor representation. $\mathbf{W}$ is a shared linear transformation, $\odot$ denotes element-wise multiplication, and $\sigma(\cdot)$ is a nonlinear activation function such as ReLU.

We then feed $\mathbf{\hat{h}}$ through a two-layer MLP to perform a binary classification (relevant vs.\ irrelevant). Formally,
\begin{equation}
\label{eq:mlp_classification}
\hat{\mathbf{y}} = \mathrm{Softmax}\bigl(\mathbf{W}_2 \,\sigma(\mathbf{W}_1 \,\mathbf{\hat{h}} + \mathbf{b}_1) + \mathbf{b}_2\bigr),
\end{equation}
where $\mathbf{W}_1, \mathbf{W}_2$ are weight matrices, $\mathbf{b}_1, \mathbf{b}_2$ are biases. 
Then, the probability of the “relevant” class in the Softmax output is used as the relevance score:
\begin{equation}
    score = \hat{\mathbf{y}}[1] \in (0,1).
\end{equation}
The top-$k$ entities with the highest scores are selected as supporting evidence.

The model is trained using cross-entropy loss with one-hot labels:
\begin{equation}
L = -\sum_{i=1}^{2} \mathbf{y_i} \log(\mathbf{\hat{y}_i}).
\end{equation}
where $y_i$ and $\hat{y}_i$ denote the ground-truth label and predicted probability for class $i$, respectively.

\begin{figure}[t]
    \centering
    \includegraphics[width=0.75\linewidth]{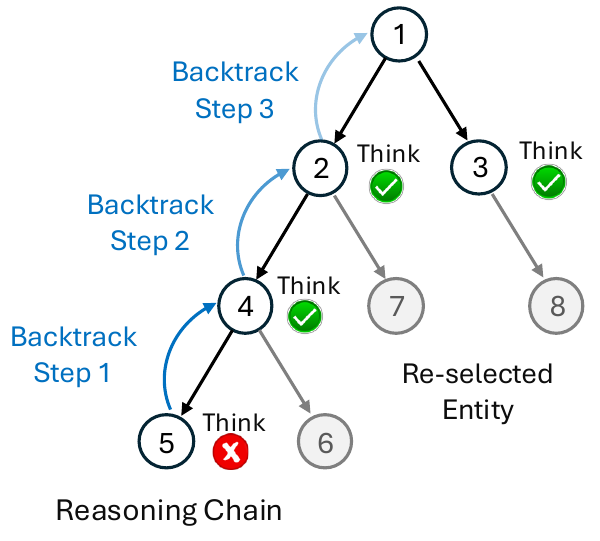}
    \caption{Backtrack for error correction.}
    \label{fig:backtrack}
\end{figure}

\subsubsection{Error Correction}

To prevent error propagation in the reasoning chain, we enable the LLM to detect and correct reasoning errors during the process. To teach the LLM how to identify errors, we intentionally introduce errors into correct reasoning chains and then revise them back to the correct form. Then we use these “error–correction” pairs to fine-tune the LLM. The detailed procedure is provided in the Appendix.

To further support error correction, we introduce a parameter \textit{Backtrack} as an input to the retrieval tool. This parameter is set to \texttt{False} by default, indicating no correction is needed. When an error is detected, the parameter is switched to \texttt{True}, triggering a backtracking process.
The backtracking relies on the previously traversed retrieval path, enabling the model to reselect candidate entities from the most semantically relevant context and continue reasoning.

An example of this process is illustrated in Figure~4. Given a reasoning path $1 \rightarrow 2 \rightarrow 4 \rightarrow 5$, if the LLM determines that entity 5 cannot support further reasoning, it backtracks to its parent entity $4$ and attempts to select an alternative entity. If no suitable candidate is found, the process continues backtracking to higher-level entities (e.g., entity 2) until a new valid reasoning path is identified or the initial topic entity is reached.

\subsection{RL-Based Enhancement}

Given the discrete nature of the reasoning process with multiple decisions and tool calls, we adopt reinforcement learning (RL) to improve the LLM’s overall reasoning performance. 
We design three reward functions to encourage correct tool usage and accurate answers.

The first is the \textit{Retrieval Reward}, which encourages the model to acquire knowledge through external retrieval rather than relying solely on its own knowledge:
\begin{equation}
R_{\text{search}} = \min(0.5 \cdot n, 0.8),
\end{equation}
where $n$ denotes the number of retrieval actions. The reward is capped at 0.8 to avoid over-encouraging excessive retrieval.

The second is the \textit{Format Reward}, which ensures that the model generates correct structural tags for reasoning, tool invocation, and answer generation, i.e., \texttt{<think>}, \texttt{<search>}, and \texttt{<answer>}:

\begin{equation}
R_{\text{format}} =
\begin{cases}
0.5 & \text{if the format is correct}, \\
0 & \text{otherwise}.
\end{cases}
\end{equation}

Finally, the \textit{Answer Reward} evaluates the correctness of the final output:

\begin{equation}
R_{\text{answer}} =
\begin{cases}
1 & \text{if the answer is correct}, \\
0 & \text{otherwise}.
\end{cases}
\end{equation}

The final reward is computed as the sum of the above components:
\begin{equation}
R_{\text{final}} = R_{\text{search}} + R_{\text{format}}+ R_{\text{answer}}  
\end{equation}

We adopt the classic Group Relative Policy Optimization (GRPO) method~\cite{shao2024deepseekmath} to optimize the LLM. The objective is to maximize the expected final reward over trajectories generated by the policy:
\begin{equation}
\theta^{*} = \arg\max_{\theta} \; \mathbb{E}{q \sim \pi_{\theta}} \big[ R_{\text{final}}(q) \big],
\end{equation}
where \(\pi_{\theta}\) denotes the policy of the LLM parameterized by \(\theta\). 

\paragraph{Hard cases Resample.}
In the first training episode, RL is applied to the full training dataset to obtain a stable initial policy. In subsequent episodes, training is restricted to difficult samples in order to improve learning efficiency and focus the optimization on challenging cases. We identify hard samples based on the next-token prediction loss computed during the supervised fine-tuning stage described in Sections~\ref{Retrieval-method} and~\ref{gnn-method}:
\begin{equation}
\mathcal{L}_{\text{sft}} = -\frac{1}{T} \sum_{t=1}^{T} \log P\left(a_t^{s} \mid a_{<t}^{s}, \mathbf{q}; \theta\right)
\end{equation}
Since a higher loss indicates a more difficult instance, we select samples satisfying $\mathcal{L}_{\text{sft}} > \epsilon$ as hard cases, where $\epsilon$ is a predefined threshold.
Furthermore, we filter these hard cases to retain only those that require multi-hop reasoning. The resulting set of hard multi-hop instances is then used in the subsequent stages of RL training.

\section{Experiments}
\subsection{Datasets}
\label{sec:datasets-experimens}
We evaluate our approach on eight widely-used datasets for KBQA, leveraging two large-scale general-purpose knowledge graphs: Freebase and WikiData. Four of the datasets are based on the Freebase knowledge graph and are standard KBQA benchmarks: ComplexWebQuestions (CWQ)~\cite{talmor2018web}, WebQuestionsSP (WebQSP)~\cite{yih2016value}, WebQuestions~\cite{berant2013semantic}, and GrailQA~\cite{gu2021beyond}.

The other four datasets are grounded in the WikiData knowledge graph. Among them, QALD10-en \cite{perevalov2022qald} is a KBQA dataset, while T-REx \cite{elsahar2018t} and Zero-Shot RE \cite{petroni2021kilt} are designed for slot filling tasks, and Creak \cite{onoe2creak} focuses on factual verification. 
Due to page limitations, the results on the four WikiData-based datasets are reported in the Appendix.

The characteristics of all datasets are provided in Table \ref{tab:datasets} in Appendix \ref{sec:additional_results}.
Following prior work~\cite{sun2024thinkongraph, zhao2024kg, xiong-etal-2024-interactive}, we use Hit@1 score as the evaluation metric for both question answering and slot filling tasks.
The implementation details are provided in Appendix \ref{sec:impl}.

\begin{table*}[ht]
\caption{
Comparison of different methods across multiple benchmarks. 
Results marked with `*' are taken directly from the corresponding original papers. 
Bold numbers indicate the \textbf{best performance}, while underlined numbers denote the \underline{second-best} (Hits@1).
}
\centering
\renewcommand{\arraystretch}{1.1}
\begin{adjustbox}{width=0.90\textwidth}
\begin{tabular}{llccccc}
\toprule
\multirow{2}{*}{\textbf{Method}} 
& \multirow{2}{*}{\textbf{LLM}} 
& \multirow{2}{*}{\textbf{Size}} 
& \multicolumn{3}{c}{\textbf{Multi-hop reasoning}} 
& \textbf{Generalization} \\
\cmidrule(lr){4-6} \cmidrule(lr){7-7}
 & & & CWQ & WebQSP & WebQuestion & GrailQA \\
\midrule

PoG~\cite{chen2024plan} 
& GPT-4 & - & 75.0$^{*}$ & 87.3$^{*}$ & \underline{84.7$^{*}$} & - \\

ToG~\cite{sun2024thinkongraph} 
& GPT-4 & - & 67.6$^{*}$ & 82.6$^{*}$ & 57.9$^{*}$ & 81.4$^{*}$ \\

LightPROF~\cite{ao2025lightprof} 
& LLaMA3 & 8B & 59.3$^{*}$ & 83.8$^{*}$ & - & - \\

RoG~\cite{luoreasoning} 
& LLaMA2 & 7B & 62.6$^{*}$ & 85.7$^{*}$ & - & - \\

KBQA-o1~\cite{luokbqa} 
& LLaMA3 & 70B & 72.0 & 88.3 & 82.5 & 72.9 \\

KG-Agent~\cite{jiang-etal-2025-kg} 
& LLaMA3 & 7B & 72.2$^{*}$ & 83.3$^{*}$ & - & \underline{86.1$^{*}$} \\

ORT~\cite{liu2025ontology} 
& DeepSeek-v3 & 671B & 72.9$^{*}$ & 89.4$^{*}$ & - & - \\

iQUEST~\cite{wang-yu-2025-iquest}  
& GPT-4o & - & 73.8$^{*}$ & 88.9$^{*}$ & 81.2$^{*}$ & 73.5$^{*}$ \\

LMP~\cite{wan2025digest}  
& GPT-4 & - & \textbf{82.2}$^{*}$ & \underline{90.0}$^{*}$ & 80.4$^{*}$ & \textbf{89.3}$^{*}$ \\

\midrule
\textbf{KG-Reasoner (Ours)} 
& Qwen3 & 30B & \underline{78.14} & \textbf{93.15} & \textbf{86.02} & 76.86 \\

\bottomrule
\end{tabular}
\end{adjustbox}
\label{tab:grouped-llm-comparison}
\end{table*}

\begin{table*}[t]
\centering
\caption{
Effect of unified reasoning and backtracking under different backbone LLMs. 
}
\small
\begin{adjustbox}{width=1\textwidth}
\begin{tabular}{llcccc}
\toprule
\textbf{Backbone LLM} & \textbf{Method} & CWQ & WebQSP & WebQuestion & GrailQA \\
\midrule

Qwen3-30B-A3B 
& Pipeline Reasoning (based on ToG) 
& 67.68 & 82.08 & 74.21 & 61.78 \\

Qwen3-30B-A3B 
& Unified Reasoning (w/o Backtracking)
& 75.25~(+7.57) & 91.86~(+9.78) & 84.76~(+10.55) & 73.63~(+11.85) \\

Qwen3-30B-A3B 
& Unified Reasoning (w/ Backtracking)
& 78.14~(+2.89) & 93.15~(+1.29) & 86.02~(+1.26) & 76.86~(+3.23) \\
\midrule

LLaMA-3.1-8B 
& Pipeline Reasoning (based on ToG) 
& 58.91 & 79.21 & 66.46 & 47.86 \\

LLaMA-3.1-8B 
& Unified Reasoning (w/o Backtracking)
& 66.18~(+7.27) & 86.45~(+7.24) & 76.63~(+10.17) & 60.70~(+12.84) \\

LLaMA-3.1-8B 
& Unified Reasoning (w/ Backtracking)
& 69.27~(+3.09) & 87.68~(+1.23) & 80.99~(+4.36) & 63.53~(+2.83) \\
\bottomrule
\end{tabular}
\end{adjustbox}
\label{tab:freebase_qwen_llama_comparison}
\end{table*}

\subsection{Main Results}
\label{sec:main-results}
We compare our approach with a range of recent and representative methods, which can be grouped into two categories: approaches based on fine-tuned open-source LLMs, including LightPROF \cite{ao2025lightprof}, RoG \cite{luokbqa}, KG-Agent \cite{jiang-etal-2025-kg}, and KBQA-o1 \cite{luokbqa}, and methods that rely on strong closed-source or extremely large LLMs, including PoG \cite{chen2024plan}, ToG \cite{sun2024thinkongraph}, iQUEST \cite{wang-yu-2025-iquest}, LMP \cite{wan2025digest}, and ORT \cite{liu2025ontology}.

The results are shown in Table \ref{tab:grouped-llm-comparison}. 
Despite using a 30B LLM, our KG-Reasoner consistently surpasses KBQA-o1, which fine-tunes a 70B model. While KG-Agent achieves competitive results by relying on a customized sub-KG extracted from the original Freebase KG, it effectively simplifies the task by artificially narrowing the search space. In contrast, our more general-purpose approach avoids such restrictive constraints.

In general, methods leveraging closed-source LLMs like GPT-4 tend to exhibit stronger performance due to their vast internal knowledge and superior reasoning depth. Among them, LMP performs best on CWQ by repeatedly invoking LLMs many times to identify second-hop entities, encode sub-graphs into natural language, and finally generate answers. This design incurs a large number of LLM calls, resulting in high computational cost and latency, which limits its practical deployment. Despite the strong capacity of GPT-4, our approach outperforms other GPT-4–based methods on three multi-hop reasoning datasets.

On the GrailQA dataset, which primarily evaluates the generalization capabilities of models, performance is typically highly correlated with parameter scale. Nevertheless, our 30B KG-Reasoner still outperforms the 70B KBQA-o1, highlighting the effectiveness of our reinforced reasoning and backtracking mechanisms. Overall, the consistent performance of our KG-Reasoner  demonstrates its robustness and efficiency in complex knowledge graph reasoning tasks.

\begin{table*}[t]
\centering
\small
\caption{
Ablation study of reinforcement learning components in KG-Reasoner using Qwen3-30B-A3B.
}
\begin{tabular}{lcccc}
\toprule
\textbf{Variant} & CWQ & WebQSP & WebQuestion & GrailQA \\
\midrule

Full Model
& 78.14 & 93.15 & 86.02 & 76.86 \\
\midrule

-- SFT Warm-Up
& 74.33 & 89.42 & 83.35 & 74.91 \\

-- Hard-Case Sampling
& 75.63 & 90.16 & 84.21 & 75.06 \\

-- Search Reward ($R_{\text{search}}$)
& 75.41 & 90.66 & 84.79 & 73.23 \\

-- Format Reward ($R_{\text{format}}$)
& 76.82 & 91.38 & 83.54 & 74.48 \\

-- Answer Reward ($R_{\text{answer}}$)
& 67.55 & 80.23 & 75.45 & 68.41 \\

\bottomrule
\end{tabular}
\label{tab:qwen_rl_ablation}
\end{table*}

\subsection{Effectiveness of the End-to-End Reasoning Framework}
We compare our end-to-end unified reasoning framework with a ToG-based pipeline and evaluate variants without and with backtracking. Experiments with Qwen3-30B-A3B and LLaMA-3.1-8B are summarized in Table~\ref{tab:freebase_qwen_llama_comparison}.

\paragraph{Benefits of Coherent Reasoning and Interleaved Retrieval.}
Comparing \emph{Pipeline Reasoning} with \emph{Unified Reasoning (w/o Backtracking)}, we observe substantial and consistent performance gains across all benchmarks. For instance, on \textsc{Qwen3-30B-A3B}, unified reasoning improves Hits@1 by $+7.57$ on CWQ and $+9.78$ on WebQSP, while achieving gains exceeding $+10$ points on both WebQuestion and GrailQA. Similar trends are observed with \textsc{LLaMA-3.1-8B}, demonstrating the robustness of the unified framework across different backbone models.
These improvements are primarily attributed to the ability of the unified framework to perform continuous reasoning with interleaved retrieval. 
Unlike fixed pipeline approaches that decompose multi-hop reasoning into independent LLM calls, our framework maintains a coherent reasoning trajectory, enabling stronger dependencies between reasoning steps. 

\paragraph{Effect of Backtracking for Error Correction.}
Introducing backtracking mechanism further enhances performance, as results of \emph{w/ Backtrack} in Table~\ref{tab:freebase_qwen_llama_comparison} shows.
During the internal thinking phase, this mechanism allows the model to autonomously identify logical dead-ends and explore alternative paths, thereby increasing the probability of locating relevant entities. Furthermore, training the model with error-correction data encourages it to self-evaluate intermediate reasoning steps.
Overall, these results demonstrate that backtracking is a crucial mechanism for robust multi-hop reasoning over knowledge graphs.

\subsection{Analysis of Reinforcement Learning Components}

To analyze the contributions of individual components in the RL process, we conduct a detailed ablation study on three key modules: SFT warm-up, hard-case sampling, and reward functions. The results are reported in Table~\ref{tab:qwen_rl_ablation}.

\paragraph{Effect of SFT Warm-Up for RL Training.}
Before RL training, we perform SFT using task-specific data to teach the LLM how to invoke retrieval tools and adapt to the reasoning pattern. As shown in Table~\ref{tab:qwen_rl_ablation}, removing the SFT warm-up stage leads to consistent performance degradation across all benchmarks. The reason is that SFT warm-up provides an effective policy initialization, preventing the model from deviating in the early stages of RL training. SFT warm-up enables more reliable optimization and results in higher-quality reasoning trajectories.

\paragraph{Efficiency Gains from Hard-Case Sampling.}
The removal of hard-case sampling (\emph{-- Hard-Case Sampling} in Table~\ref{tab:qwen_rl_ablation}) results in a obvious performance decline. The reason is that focusing on challenging multi-hop queries prevents the RL process from over-fitting to trivial samples and incentivizes the model to explore more complex reasoning trajectories. Additional analyses of training dynamics, including reward curves, response length, and retrieval frequency, are visualized in Figure~\ref{fig:three_images} (Appendix~\ref{sec:additional_results}). 

\paragraph{Ablation of Reward Functions.}
The ablation results in Table~\ref{tab:qwen_rl_ablation} show that removing any individual reward component consistently degrades performance, confirming the complementary nature of our reward design. The answer reward $R_{\text{answer}}$ has the largest impact, as it provides direct supervision from ground-truth labels and aligns the model with the primary task objective. In addition, the search reward $R_{\text{search}}$ and format reward $R_{\text{format}}$ offer effective guidance by encouraging proactive knowledge retrieval and adherence to the predefined response structure, respectively. Overall, these results highlight the importance of a multi-faceted reward design for RL-based KG reasoning.

\section{Conclusion}
In this paper, we presented KG-Reasoner, an end-to-end framework for multi-hop knowledge graph question answering that unifies retrieval and reasoning within a single long-thinking process of a reasoning-oriented LLM. By internalizing KG traversal through reinforcement learning, the model is able to perform coherent multi-step reasoning, dynamically explore reasoning paths, and backtrack from erroneous decisions. Experimental results on eight challenging benchmarks demonstrate that KG-Reasoner achieves competitive or superior performance compared to state-of-the-art methods. Our results highlight the potential of reinforcement-trained reasoning LLMs as a unified solution for complex multi-hop KBQA tasks.

\section*{Limitations}
While our approach has shown promising effectiveness, it still faces several limitations. First, improving the reasoning capability of LLMs during the "think" stage often requires reinforcement learning. Optimization algorithms such as Group Relative Policy Optimization (GRPO) involve coordinating the policy model, reference model, value model, and reward models, which incurs substantial computational overhead. In future work, we plan to investigate more resource-efficient optimization strategies.

Second, our method does not directly address the incompleteness and noise present in underlying knowledge graphs. This is a well-known challenge in KBQA. Since reasoning relies heavily on the correctness of entities and edges, missing or erroneous links can cause inference failures. As future work, we plan to explore incorporating redundant evidence and developing more robust retrieval techniques to mitigate the impact of KG imperfections.

\section*{Acknowledgment}
This work was partially funded by the Autonomous Systems and Software Program (WASP), supported by the Knut and Alice Wallenberg Foundation, and the Chalmers Artificial Intelligence Research Centre (CHAIR).

\bibliography{custom}

\appendix

\section{Implementation Details}
\label{sec:impl}
We conduct experiments using two instruction-tuned language models as backbones: LLaMA-3.1-8B-Instruct and Qwen3-30B-A3B. To mitigate the unstable cold-start phase of reinforcement learning, we first construct a set of 500 high-quality examples to teach the models how to properly invoke the knowledge retrieval tool. We randomly sample 2,000 examples from 8 datasets for RL training. The scoring model $f_r$ for the reasoning process is Llama-3.3-70B, while the model $f_a$ used to determine whether the predicted answer matches the ground truth is Llama-3.2-3B. 
In the main experiments, to ensure fair comparison, we employ GPT-4o within the KG retrieval tool to assist in selecting relevant entities. We restrict its use strictly to the KG retrieval module, while the reasoning process relies solely on LLaMA-3.1-8B or Qwen-2.5-7B. In all other experiments, no external large language models such as GPT-4o are introduced, in order to better isolate and evaluate the capabilities of our proposed framework.
Starting from the second epoch, we apply a resampling strategy to dynamically filter out trivial questions and retain more informative ones for continued training.
The training is performed on 8 NVIDIA A100 80G GPUs in total. For each input query, we generate 16 outputs (rollouts). We train for 2 epochs with a batch size of 16 and a learning rate of $1\text{e}{-6}$. The rollout temperature is set to 1, the PPO clip ratio is 0.2, and the KL divergence penalty coefficient is $1\text{e}{-5}$. 

\section{Additional Experimental Details}
\label{sec:additional_results}
\begin{table*}[ht]
\centering
\caption{Overview of datasets for KBQA and related tasks}
\label{tab:datasets}
\resizebox{\textwidth}{!}{
\begin{tabular}{@{}lllllllll@{}}
\toprule
\textbf{Dataset} & \textbf{KG} & \textbf{Task} & \textbf{Train} & \textbf{Test} & \textbf{Complexity} & \textbf{Reasoning} & \textbf{Annotation} & \textbf{Source} \\
\midrule
CWQ           & Freebase & KBQA          & 27,689          & 3,531 & Complex Qs           & Multi-hop                 & SPARQL, Ans.           & Web-derived \\
WebQSP        & Freebase & KBQA          & 3,098           & 1,639 & Moderate Qs          & Single-hop (most)         & SPARQL, Ans.           & Web-derived \\
WebQuestion   & Freebase & KBQA          & 3,778           & 2,032 & Simple–Moderate Qs   & Single-hop (most)         & Ans. only              & Web-derived \\
GrailQA       & Freebase & KBQA          & 35,138          & 1,000 & Complex (varied)     & Multi-hop                 & Logic forms, Ans.      & Human-controlled \\
\midrule
QALD10-en     & Wikidata & KBQA          & 412             & 394   & Complex Qs           & Multi-hop                 & SPARQL, Ans.           & Human experts \\
T-REx         & Wikidata & Slot Filling  & $\sim$11M       & 5,000 & Text (abstracts)     & N/A (extraction)          & Triples + text         & Wikidata \\
Zero-Shot RE  & Wikidata & Slot Filling  & $\sim$860k      & 3,724 & Text (sentences)     & N/A (inference)           & Text + rel. defs.      & Wikidata \\
Creak         & Wikidata & Fact Verify   & 10,176          & 1,371 & Complex claims       & Factual and CS infer.      & Labels, expls.         & Human-authored \\
\bottomrule
\end{tabular}
}
\end{table*}

\begin{figure*}[ht]
  \centering
  \begin{subfigure}{0.32\textwidth}
    \includegraphics[width=\linewidth]{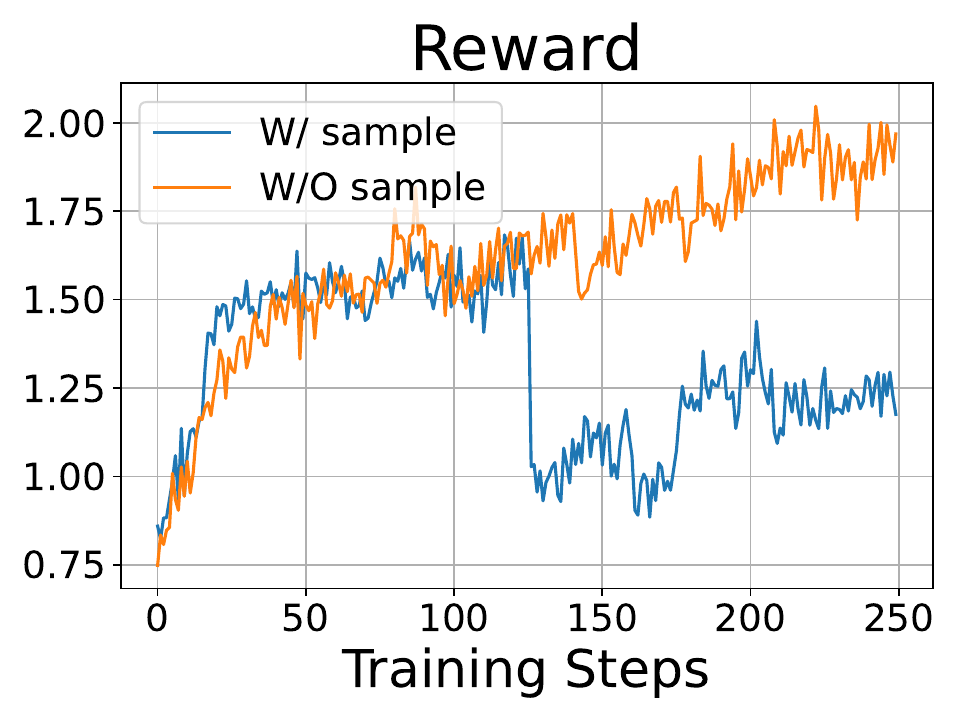} 
  \end{subfigure}
  \begin{subfigure}{0.32\textwidth}
    \includegraphics[width=\linewidth]{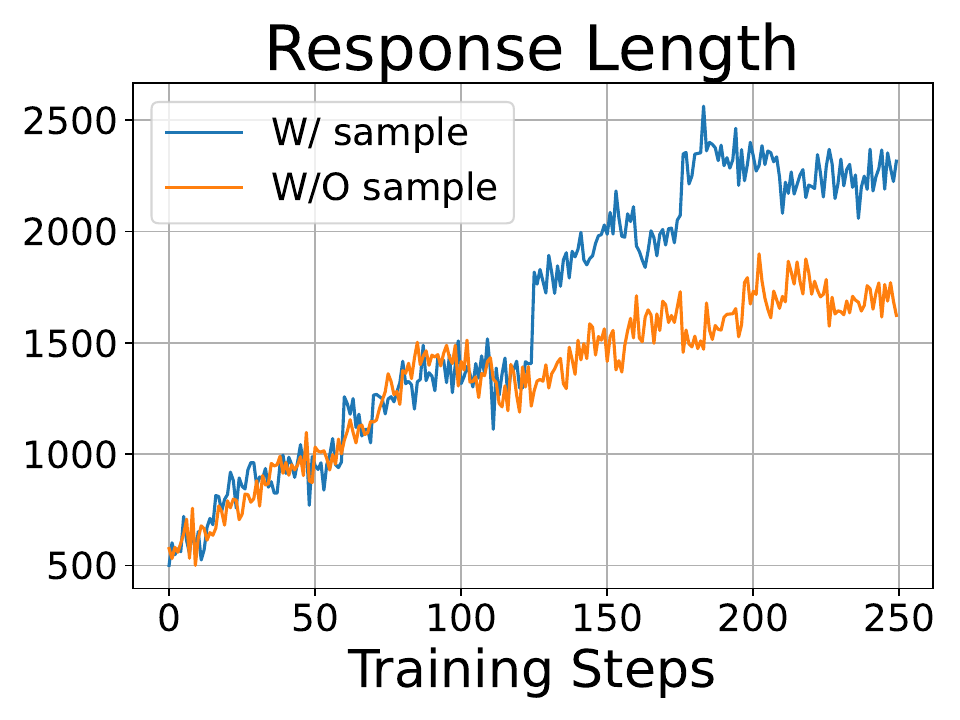}
  \end{subfigure}
  \begin{subfigure}{0.32\textwidth}
    \includegraphics[width=\linewidth]{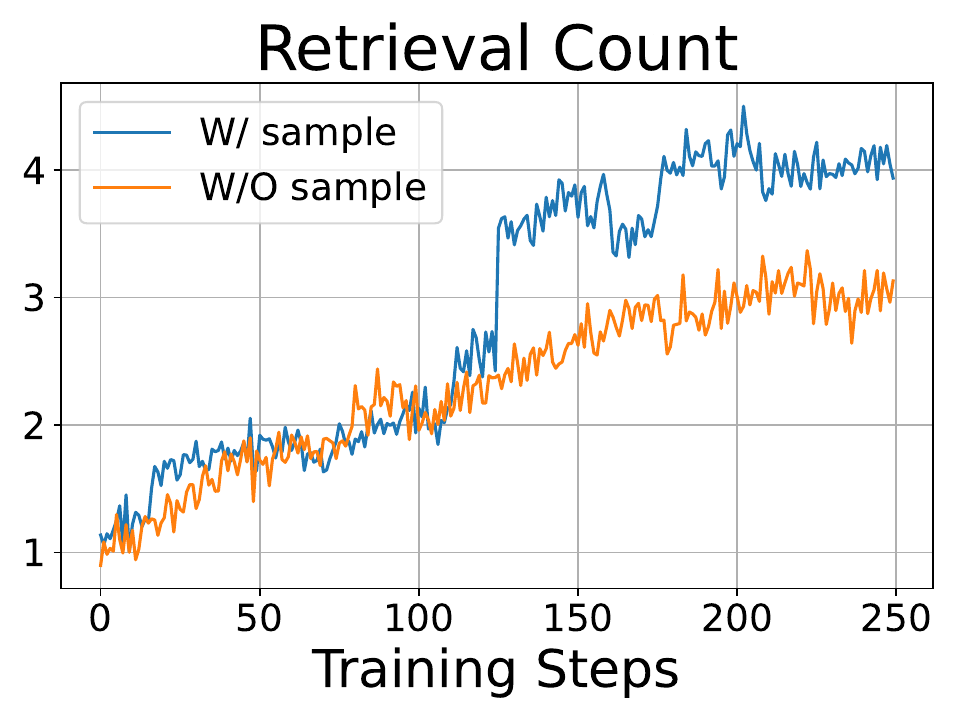}
  \end{subfigure}
  \caption{The RL training process under two settings: with and without hard case sampling. The figure shows how reward, response length, and retrieval count change over training steps.}
  \label{fig:three_images}
\end{figure*}

\subsection{Impact of Graph-Based Context Aggregation.}
We evaluate the effectiveness of the proposed graph-based context aggregation module (Section~\ref{sec:gnn}) across two backbone LLMs, with results summarized in Table~\ref{tab:gat}. Incorporating the GNN module consistently improves performance across all benchmarks. By aggregating information from neighboring entities, the module enriches entity representations beyond isolated semantics, leading to more accurate retrieval. Overall, these results demonstrate that graph-based context aggregation provides a robust and easily pluggable enhancement that effectively leverages local KG structure to improve retrieval performance.

\begin{table*}[t]
\centering
\caption{
Impact of GNN-based context aggregation in KG-Reasoner across different backbone LLMs.
}
\small
\begin{adjustbox}{width=1\textwidth}
\begin{tabular}{llcccc}
\toprule
\textbf{Backbone LLM} & \textbf{Method} & CWQ & WebQSP & WebQuestion & GrailQA \\
\midrule

Qwen3-30B-A3B 
& KG-Reasoner (w/o Context Aggregation)
& 75.62 & 90.90 & 81.68 & 71.87 \\

Qwen3-30B-A3B 
& KG-Reasoner (w/ Context Aggregation)
& 78.14~(+2.52) & 93.15~(+2.25) & 86.02~(+4.34) & 76.86~(+4.99) \\
\midrule

LLaMA-3.1-8B 
& KG-Reasoner (w/o Context Aggregation)
& 64.88 & 84.37 & 70.45 & 57.70 \\

LLaMA-3.1-8B 
& KG-Reasoner (w/ Context Aggregation)
& 69.27~(+4.39) & 87.68~(+3.31) & 80.99~(+10.54) & 63.53~(+5.83) \\
\bottomrule
\end{tabular}
\end{adjustbox}
\label{tab:gat}
\end{table*}


\section{Prompts}
\label{sec:prompts}

\begin{figure*}[t]
\centering
\begin{minipage}{\textwidth}
\begin{tcolorbox}[
  colback=gray!5!white,
  colframe=green!30!black,
  coltitle=white,
  fonttitle=\bfseries,
  title=Prompt for Constructing Warm Up Data,
  colbacktitle=green!30!black,
  sharp corners=south,
  enhanced,
  boxrule=0.5pt,
  drop shadow=black!20,
  arc=2mm,
  width=\textwidth
]
You are a knowledgeable assistant that answers user questions through explicit reasoning and, when necessary, knowledge graph searches.

Instructions:

1. The assistant first thinks through the reasoning process before providing the final answer.

2. The reasoning process must be enclosed in `<think> ... </think>` tags.

3. The final answer must be enclosed in `<answer> ... </answer>` tags.

4. If the assistant lacks specific knowledge during reasoning, it may query a knowledge graph by issuing a search request using the format:
   `<search> [ENTITY] </search>`

   * Only one entity is allowed per search.
   * The entity must be related to the question (e.g., a topic entity or one present in previously retrieved triples).
   
5. The system will respond with relevant knowledge in the format:
   `<searched\_triples> (subject, predicate, object) </searched\_triples>`
   
6. The assistant must incorporate any retrieved triples into its reasoning process.

Example:

User: What timezone is Utah in?
Topic Entity: Utah

Assistant:

```
<think>  
I am unsure about the timezone of Utah.  
I will perform a search to retrieve relevant information.  
<search>Utah</search>  
<searched\_triples> (Utah, timeZone, Mountain Time Zone) </searched\_triples>  
Based on the search results, I found that Utah is in the Mountain Time Zone.  
</think>  
<answer>Mountain Time Zone</answer>  
```

Now begin answering the question.

User: [INPUT\_QUESTION]
Topic Entity: [TOPIC\_ENTITY]
\end{tcolorbox}
\end{minipage}
\end{figure*}

\begin{figure*}[t]
\centering
\begin{minipage}{\textwidth}
\begin{tcolorbox}[
  colback=gray!5!white,
  colframe=green!30!black,
  coltitle=white,
  fonttitle=\bfseries,
  title=System Prompt for LLMs,
  colbacktitle=green!30!black,
  sharp corners=south,
  enhanced,
  boxrule=0.5pt,
  drop shadow=black!20,
  arc=2mm,
  width=\textwidth
]
The User asks a question, and the Assistant solves it.
The Assistant first thinks about the reasoning process in the mind and then provides the User with the final answer.
The output format of reasoning process and final answer are enclosed within <think> </think> and <answer> </answer> tags, respectively, i.e., "<think> reasoning process here </think>~
<answer> final answer here </answer>".
During the thinking process, the Assistant can perform searching for uncertain knowledge from a knowledge graph if necessary with the format of "<search> search query (only one entity) </search>". The entity must be involved in the searched triples or topic entities.
Then, the system will provide the Assistant with helpful information with the format of "<triples> ...search results... </triples>".

User: [INPUT\_QUESTION]
~~Topic Entity: [TOPIC\_ENTITY]

Assistant:
\end{tcolorbox}
\end{minipage}
\end{figure*}

\begin{table*}[ht]
\caption{
Full Results (including ablation study) on the datasets using WikiData.
}
\centering
\renewcommand{\arraystretch}{1.1}
\begin{adjustbox}{width=0.95\textwidth}
\begin{tabular}{lccccc}
\toprule
\multirow{2}{*}{\textbf{Method}} & \multirow{2}{*}{\textbf{Size}} & \multicolumn{4}{c}{\textbf{WikiData}} \\
\cmidrule(lr){3-6}
 & & QALD10-en & T-REx & Zero-Shot RE & Creak \\
\midrule

\rowcolor{gray!10} \multicolumn{6}{l}{\textbf{LLM w/o KG}} \\
Qwen-2.5-7B & 7B & 41.88 & 31.15 & 7.84 & 73.26 \\
LLaMA-3.1-8B & 8B & 40.25 & 23.00 & 12.54 & 75.80 \\
LLaMA-3.3-70B & 70B & 56.00 & 20.12 & 18.55 & 83.72 \\
DeepSeek-R1-Distill-Llama-70B & 70B & 43.10 & 34.21 & 22.27 & 79.10 \\
GPT-4o-mini & – & 51.98 & 26.90 & 18.85 & 83.72 \\
GPT-4o & – & 56.20 & 44.46 & 48.20 & 90.70 \\
\midrule
\rowcolor{gray!10}\multicolumn{6}{l}{\textbf{KG + LLMs w/o Fine-Tuning}} \\
Qwen-2.5-7B + KG & 7B & 56.54 & 64.21 & 70.32 & 79.04 \\
LLaMA-3.1-8B + KG & 8B & 55.73 & 65.80 & 68.66 & 80.50 \\
LLaMA-3.3-70B + KG & 70B & 71.78 & 65.42 & {80.42} & 87.04 \\
DeepSeek-R1-Distill-Llama-70B + KG & 70B & 66.10 & 72.04 & 78.04 & {91.20} \\
GPT-4o-mini + KG & – & \underline{72.86} & 69.70 & 75.12 & 90.20 \\
ToG with GPT 3.5-Turbo~\cite{sun2024thinkongraph} & – & 50.2$^{*}$ & \underline{76.8$^{*}$} & {88.0$^{*}$} & {91.2$^{*}$} \\
ToG with GPT 4~\cite{sun2024thinkongraph} & – & 53.8$^{*}$ & {77.1$^{*}$} & {88.3$^{*}$} & \underline{95.6$^{*}$} \\
ToG-2.0 (ICL)(GPT-3.5-turbo) ~\cite{mathink} & -  & 54.1$^{*}$ & - & \underline{91.0$^{*}$} & 93.5$^{*}$ \\
\midrule
\rowcolor{gray!10}\multicolumn{6}{l}{\textbf{KG + LLM with Fine-Tuning (LLM Backbone Fine-Tuning)}} \\
Qwen-2.5-7B (SFT) + KG & 7B & 65.18 & 68.77 & 71.46 & 83.43 \\
LLaMA-3.1-8B (SFT) + KG & 8B & 59.63 & 70.23 & 64.08 & 84.47 \\
\textbf{KG-Reasoner with Qwen-2.5-7B (Ours)} & 7B & \textbf{83.73} & \textbf{83.68} & \textbf{91.43} & \textbf{97.45} \\

\bottomrule
\end{tabular}
\end{adjustbox}
\label{tab:grouped-llm-comparison-wikidata}
\end{table*}

\begin{table*}[ht]
\caption{
Full results (including ablation study) on four datasets using Freebase.
}
\centering
\renewcommand{\arraystretch}{1.1}
\begin{adjustbox}{width=0.78\textwidth}
\begin{tabular}{lccccc}
\toprule
\multirow{2}{*}{\textbf{Method}} & \multirow{2}{*}{\textbf{Size}} & \multicolumn{4}{c}{\textbf{Freebase}} \\
\cmidrule(lr){3-6}
 & & CWQ & WebQSP & WebQuestion & GrailQA \\
\midrule

\rowcolor{gray!10} \multicolumn{6}{l}{\textbf{LLM w/o KG}} \\
Qwen-2.5-7B & 7B & 31.25 & 46.97 & 44.23 & 29.53 \\
LLaMA-3.1-8B & 8B & 32.33 & 45.07 & 45.88 & 28.35 \\
LLaMA-3.3-70B & 70B & 37.20 & 71.12 & 59.73 & 33.79 \\
DeepSeek-R1-Distill-Llama-70B & 70B & 31.92 & 77.43 & 68.84 & 31.70 \\
GPT-4o-mini & – & 42.32 & 65.97 & 57.26 & 36.22 \\
GPT-4o & – & 41.77 & 72.55 & 64.79 & 35.01 \\
\midrule

\rowcolor{gray!10} \multicolumn{6}{l}{\textbf{KG + LLMs w/o Fine-Tuning}} \\
Qwen-2.5-7B + KG & 7B & 44.82 & 72.60 & 56.33 & 41.10 \\
LLaMA-3.1-8B + KG & 8B & 45.64 & 71.32 & 57.05 & 40.40 \\
LLaMA-3.3-70B + KG & 70B & 44.00 & 81.90 & 72.60 & 57.70 \\
DeepSeek-R1-Distill-Llama-70B + KG & 70B & 52.38 & 81.34 & 75.80 & 59.02 \\
GPT-4o-mini + KG & – & 54.35 & 84.40 & {81.02} & 60.00 \\
PoG with GPT-4~\cite{chen2024plan} & - & 75.0$^{*}$ & \underline{87.3$^{*}$} & \underline{84.7$^{*}$} & - \\
ToG with GPT 4~\cite{sun2024thinkongraph} & – & 67.6$^{*}$ & 82.6$^{*}$ & 57.9$^{*}$ & \underline{81.4$^{*}$} \\
Readi with GPT-4~\cite{cheng2024call} & - & 67.0$^{*}$ & 78.7$^{*}$ & - & - \\
\midrule

\rowcolor{gray!10} \multicolumn{6}{l}{\textbf{KG + LLM with Fine-Tuning (Small Models Fine-Tuning)}} \\
KG-CoT with GPT 4~\cite{zhao2024kg} & – & 62.3$^{*}$ & 84.9$^{*}$ & 68.0$^{*}$ & - \\
Chain-of-Question~\cite{yixing2024chain} & - & \textbf{78.8$^{*}$} & 78.10$^{*}$ & - & - \\
FlexKBQA~\cite{li2024flexkbqa} & - & 46.20$^{*}$ & - & 68.90$^{*}$ & - \\
LightPROF (LLaMA3-8B)~\cite{ao2025lightprof} & 8B & 59.3$^{*}$ & 83.8$^{*}$ & - & - \\
\midrule

\rowcolor{gray!10} \multicolumn{6}{l}{\textbf{KG + LLM with Fine-Tuning (LLM Backbone Fine-Tuning)}} \\
Qwen-2.5-7B (SFT) + KG & 7B & 51.84 & 74.80 & 61.42 & 46.18 \\
LLaMA-3.1-8B (SFT) + KG & 8B & 47.40 & 72.98 & 60.00 & 47.70 \\
G-Retriever~\cite{he2024g} & 7B & - & 73.79$^{*}$ & - & - \\
Interactive-KBQA (7B)~\cite{xiong-etal-2024-interactive} & 7B & 39.9$^{*}$ & 43.57$^{*}$ & - & - \\
Interactive-KBQA (13B)~\cite{xiong-etal-2024-interactive} & 13B & 42.5$^{*}$ & 54.86$^{*}$ & - & - \\
RoG~\cite{luoreasoning} & 7B & 62.6$^{*}$ & 85.7$^{*}$ & - & - \\
KBQA-o1~\cite{luokbqa} & - & 67.0$^{*}$ & - & {81.6$^{*}$} & - \\
KG-Agent~\cite{jiang-etal-2025-kg} & - & 72.2$^{*}$ & 83.3$^{*}$ & - & \textbf{86.1$^{*}$} \\
\textbf{KG-Reasoner (Ours)} & 30B & \underline{78.14} & \textbf{93.15} & \textbf{86.02} & 76.86 \\

\bottomrule
\end{tabular}
\end{adjustbox}
\label{tab:grouped-llm-comparison}
\end{table*}

\begin{table*}[t]
\centering
\small
\caption{Performance comparison of different reasoning and training strategies on WikiData benchmarks.}
\begin{tabular}{lcccc}
\toprule
Method & QALD10-en & T-REx & Zero-Shot RE & Creek \\
\midrule
Qwen3-30B-A3B &
55.93 & 19.68 & 17.89 & 83.35 \\

Qwen3-30B-A3B + KG rag &
71.86 & 65.39 & 80.43 & 86.83 \\

Qwen3-30B-A3B + fixed pipeline &
73.06 & 70.48 & 83.71 & 87.67 \\

Qwen3-30B-A3B + RL &
81.65 & 80.32 & 88.05 & 93.76 \\

Qwen3-30B-A3B + RL + GAT &
83.73 & 83.68 & 91.43 & 97.45 \\
\midrule
LLaMA-3.1-8B &
40.25 & 23.04 & 12.54 & 75.80 \\

LLaMA-3.1-8B + KG rag &
55.73 & 65.80 & 68.66 & 80.50 \\

LLaMA-3.1-8B + fixed pipeline &
60.90 & 69.59 & 75.48 & 85.66 \\

LLaMA-3.1-8B + RL &
67.39 & 74.07 & 79.98 & 90.03 \\

LLaMA-3.1-8B + RL + GAT &
73.43 & 78.12 & 85.04 & 93.04 \\
\bottomrule
\end{tabular}
\label{tab:wikidata_qwen_llama_comparison}
\end{table*}

\end{document}